\documentclass{article}

\usepackage{amsmath}
\usepackage{amssymb}
\usepackage{amsthm}
\usepackage{placeins}
\usepackage{wrapfig}
\usepackage{nicefrac}

\usepackage{hyperref}

\newcommand{\E}[2]{\operatorname{\mathbb{E}}_{#1}\left[#2\right]}

\newcommand{\density}{p}

\newcommand{\kl}[2]{\mathrm{D_{KL}}\left(#1\;\middle\|\;#2\right)}

\newcommand{\ent}{\mathcal{H}}

\newcommand{\voidarg}{{\,\cdot\,}}

\newcommand{\state}{\mathbf{s}}

\newcommand{\st}{{\state_t}}

\newcommand{\stp}{{\state_{t+1}}}

\newcommand{\pdyn}{\density}

\newcommand{\action}{\mathbf{a}}

\newcommand{\at}{{\action_t}}

\newcommand{\reward}{r}

\newcommand{\V}{V}

\newcommand{\Q}{Q}

\newcommand{\policy}{\pi}

\newcommand{\params}{\theta}

\newcommand{\pparams}{{\phi}}   
\newcommand{\vparams}{{\psi}}   
\newcommand{\vtargetparams}{{\bar\psi}}   

\newcommand{\gauss}{\mathcal{N}}

\newcommand{\discount}{\gamma}

\usepackage[preprint, nonatbib]{neurips_2019}

\usepackage[utf8]{inputenc} 
\usepackage[T1]{fontenc}    
\usepackage{hyperref}       
\usepackage{url}            
\usepackage{booktabs}       
\usepackage{amsfonts}       
\usepackage{nicefrac}       
\usepackage{microtype}      

\usepackage{graphicx}
\graphicspath{ {./figures/} }
\usepackage{subcaption}
\usepackage{algorithm}
\usepackage{algorithmicx,algpseudocode}
\usepackage{amsmath,amssymb}
\title{Boosting Soft Actor-Critic: Emphasizing Recent Experience without Forgetting the Past}

\author{%
  Che Wang\\
  New York University/NYU Shanghai\\
  \texttt{cw1681@nyu.edu}\\
  \And
  Keith Ross\\
  NYU Shanghai/New York University\\
  \texttt{keithwross@nyu.edu}\\
}
\begin{document}
\maketitle

\begin{abstract}
Soft Actor-Critic (SAC) \cite{haarnoja2018soft, haarnoja2018softapps} is an off-policy actor-critic deep reinforcement learning (DRL) algorithm based on maximum entropy reinforcement learning. By combining off-policy updates with an actor-critic formulation, SAC achieves state-of-the-art performance on a range of continuous-action benchmark tasks, outperforming prior on-policy and off-policy methods. 
The off-policy method employed by SAC samples data uniformly from past experience when performing parameter updates. We propose Emphasizing Recent Experience (ERE), a simple but powerful off-policy sampling technique, which emphasizes recently observed data while not forgetting the past. The ERE algorithm samples more aggressively from recent experience, and also orders the updates to ensure that updates from old data do not overwrite updates from new data. We compare vanilla SAC and SAC+ERE, and show that ERE is more sample efficient than vanilla SAC for continuous-action Mujoco tasks \cite{todorov2012mujoco}. We also consider combining SAC with Priority Experience Replay (PER) \cite{schaul2015prioritized}, a scheme originally proposed for deep Q-learning which prioritizes the data based on temporal-difference (TD) error. We show that SAC+PER can marginally improve the sample efficiency performance of SAC, but much less so than SAC+ERE. Finally, we propose an algorithm which integrates ERE and PER and show that this hybrid algorithm can give the best results for some of the Mujoco tasks. 
\end{abstract}

\section{Introduction}
Soft Actor-Critic \cite{haarnoja2018soft, haarnoja2018softapps} is an off-policy actor-critic deep reinforcement learning (DRL) algorithm based on maximum entropy reinforcement learning. By combining off-policy updates with an actor-critic formulation, SAC achieves state-of-the-art performance on a range of continuous-action benchmark tasks, outperforming prior on-policy and off-policy methods. Furthermore, SAC has been shown to be relatively robust, achieving similar performance across different initial random seeds. 


SAC is an off-policy method which uses a buffer to store past experience for experience replay \cite{lin1992experiencereplay}. SAC samples data uniformly from the buffer when performing parameter updates. A uniform sampling scheme implicitly assumes that data in the replay buffer are of equal importance.
However, intuitively it is more important to build relatively accurate function approximators in regions of the state and action spaces for which the current policy is likely to operate. At the same time, it is important that the function approximators also be reasonably accurate in other regions where the policy may visit with lower probability.

To address this problem, we propose Emphasizing Recent Experience (ERE), a simple but powerful off-policy sampling technique, which emphasizes recently observed data while not forgetting the past. When performing updates, the ERE algorithm samples more aggressively from recent experience, and also orders the updates to ensure that updates from old data do not overwrite updates from new data. We compare vanilla SAC and SAC+ERE, and show that ERE provides significant performance improvements over SAC in terms of sample efficiency for continuous-action Mujoco tasks. It provides this improvement without degrading the excellent robustness of SAC. 

We also consider combining SAC with Prioritized Experience Relay (PER) \cite{schaul2015prioritized}, a scheme originally proposed for deep Q-learning which prioritizes the data based on the temporal-difference (TD) error. We show that SAC+PER can marginally improve the sample efficiency performance of SAC, but much less so than SAC+ERE. We also compare the programming and computational complexity of ERE with PER, and show that ERE is easier to implement, with no special data structure required, and fewer hyper-parameters, which are also easier to optimize. Finally, we propose an algorithm which integrates ERE and PER and show that it gives the best results for some environments.  

\section{Overview of Experience Replay and Related Work}
Experience replay \cite{lin1992experiencereplay} is a simple yet powerful method for enhancing the performance of an off-policy DRL algorithm. Experience replay stores past experience in a replay buffer and reuses this past data when making updates. Experience replay achieved great successes in Deep Q-Networks (DQN) \cite{mnih2013dqn, mnih2015dqn}. In DQN, a large buffer of size 1 million is used to store past experience, and the algorithm samples data uniformly from this large buffer for each mini-batch update. 

Experience replay schemes alternate between two phases: a data collection phase and a parameter update phase. In the data collection phase, the current policy interacts with the environment to generate transitions, which are added to a replay buffer $D$. Each data point is a tuple $(s,a,r,s')$, where $s$ is the current state, $a$ is the action taken, $r$ is the resulting reward, and $s'$ is the subsequent state. The replay buffer is fixed to a finite size (e.g., one million data points) so that very old data is dropped from the buffer. During the parameter update phase, the parameters of the neural networks are updated with samples drawn from the replay buffer $D$. Typically, this phase consists of several iterations, with each iteration drawing a mini-batch of data from $D$ and updating the parameters using the mini-batch. At the end of these iterations, the new parameters provide a new policy.

When doing mini-batch update with data from the replay buffer, a straightforward method is to simply sample uniformly from the buffer. Many other sampling methods have been proposed in the past, and one of the most well-known methods is prioritized experience replay (PER) \cite{schaul2015prioritized}. PER uses the absolute TD-error of a data point as the measure for priority, and data points with higher priority will have a higher chance of being sampled. This method has been tested on DQN \cite{mnih2015dqn} and double DQN (DDQN) \cite{van2016ddqn}, and results show significant improvement over using uniform sampling. 
PER has been combined with the dueling network architecture in \cite{wang2015dueling}, with an ensemble of recurrent DQN in \cite{schulze2018vizdoom}, and PER is one of six crucial components in Rainbow \cite{hessel2018rainbow}, which achieves state-of-the-art on the Atari game environments. PER has also been successfully applied to other algorithms such as DDPG \cite{hou2017ddpgper} and can be implemented in a distributed manner \cite{horgan2018distributed}.

There are other methods proposed to make better use of the replay buffer. In Sample Efficient Actor-Critic with Experience Replay (ACER), the algorithm has an on-policy part and an off-policy part, with a hyper-parameter controlling the ratio of off-policy updates to on-policy updates \cite{wang2016acer}. The RACER algorithm \cite{novati2018remember} selectively removes data points from the buffer, based on the degree of "off-policyness" which is measured by their importance sampling weight, bringing improvement to DDPG \cite{lillicrap2015ddpg}, NAF \cite{gu2016naf} and PPO \cite{schulman2017proximal}.  
In \cite{de2015replaydatabase}, replay buffers of different sizes were tested on DDPG, and result shows that a large enough buffer with enough data diversity can lead to better performance. Finally, with Hindsight Experience Replay (HER)\cite{andrychowicz2017her}, 
priority can be given to trajectories with lower density estimation\cite{zhao2019curiosity} to tackle multi-goal, sparse reward environments.

To our knowledge, this is the first paper that considers non-uniform data sampling techniques for SAC, and also the first paper to consider the ERE scheme for off-policy DRL algorithms.


\section{Emphasizing Recent Experience}

In this section we first give a brief review of the SAC algorithm. We then propose three SAC variants for enhancing experience replay. Pseudo-code for each variant can be found in the Appendix.
\subsection{Soft Actor-Critic Algorithm}

Soft Actor-Critic (SAC) \cite{haarnoja2018soft} is a model-free off-policy  deep reinforcement learning (DRL) algorithm based on maximum entropy reinforcement learning. By combining off-policy updates with an actor-critic formulation, SAC achieves state-of-the-art performance on a range of continuous-action benchmark tasks, outperforming prior on-policy and off-policy methods, including proximal policy optimization (PPO) \cite{schulman2017proximal}, deep determinisitc policy gradient (DDPG) \cite{lillicrap2015ddpg}, soft Q-learning \cite{haarnoja2017sql}, twin delayed determinisitc policy gradient (TD3) \cite{fujimoto2018td3}, and trust region path consistency learning (Trust-PCL) \cite{nachum2017trustpcl}. The experimental results show that SAC consistently outperforms the other RL algorithms for continuous-action benchmarks, both in terms of learning speed and robustness \cite{haarnoja2018soft}.

Here we give a brief summary of Soft Actor-Critic (SAC); for more details please refer to the SAC paper \cite{haarnoja2018soft}. SAC tries to maximize the expected sum of rewards and the entropy of a policy $\policy$: 
\begin{align}
\label{eq:maxent_objective}
J(\policy)  = \sum_{t=0}^{T} \E{(\st, \at) \sim \rho_\policy}{\reward(\st,\at) + \alpha\ent(\policy(\voidarg|\st))}.
\end{align}
Here $\rho_\policy$ is the state-action marginals of the trajectory distribution induced by $\policy$. The hyper-parameter $\alpha$ balances exploitation and exploration, and affects the stochasticity of the optimal policy\cite{haarnoja2018soft}.

SAC consists of five networks: a policy network $\phi$ that takes in the state and outputs the mean and standard deviation of an action distribution; two Q-networks $\theta_1$, $\theta_2$ to estimate the value of state-action pairs; a value network $\psi$ that estimates the value of a state; and a target value network $\bar{\psi}$ which is simply an exponentially moving average of the value network $\psi$. Since SAC is an off-policy scheme employing experience replay, it alternates between a data collection phase using the current policy, and a parameter update phase, where mini-batches of data are uniformly sampled from the replay buffer to perform updates of the parameters. In the original SAC implementation, only one sample (one interaction with the environment) is collected during the data collection phase, and one mini-batch update is performed during the update phase.  
In our implementation, we first collect data for an episode until it terminates, either because of a bad action, or reaching 1000 timesteps; we then set the number of mini-batch updates to be the same as the length of the episode. Both of these implementations give almost the same sample-efficiency and robustness performance. 

In SAC, the maximum entropy formulation is a critical component that enhances its exploration and robustness \cite{ziebart2008maximum, haarnoja2017sql}. 
In a recently updated version of SAC \cite{haarnoja2018softapps}, the entropy term $\alpha$ is learned and adapted for each environment. The new version performs better than the earlier version in many but not all environments. In this paper, we use the original and simpler SAC \cite{haarnoja2018soft} for constructing new variants using non-uniform sampling. 

\subsection{Soft Actor-Critic with Emphasizing Recent Experience}
In this section we propose SAC with Emphasizing Recent Experience (SAC+ERE), a simple yet powerful method for replaying experience. The core idea is that during the parameter update phase, the first mini-batch is sampled from all the data in the replay buffer, then for each subsequent mini-batch we gradually reduce our range of sampling to sample more aggressively from more recent data points. There are two key points of this scheme: $(i)$ we sample more recent data with higher frequency; $(ii)$ we arrange the updates so that updates with older data do not overwrite the updates with the fresher data. 


Specifically, assume that in the current update phase we are to make $K$ mini-batch updates. Let $N$ be the max size of the replay buffer. Then for the $k$th update, $1 \leq k \leq K$, we sample uniformly from the most recent $c_k$ data points, where 

\begin{equation}
c_k = \max \{ N \cdot \eta^{k\frac{1000}{K}}, \; c_{min} \}
\end{equation}

where $\eta \in (0,1]$ is a hyperparameter that determines how much emphasis we put on recent data. When $\eta=1$ this is equivalent to uniform sampling. In our experiments we found $\eta=0.996$ to be a good value for all environments. When $\eta < 1$, $c_k$ decreases as we perform each update. We set $c_{min}$ as the minimum allowable value of $c_k$. This can help prevent sampling from a very small amount of recent data, which may cause overfitting. We used this formulation here instead of just $c_k = N \cdot \eta^{k}$ because the length of an episode can vary greatly depending on the environment, and it can be beneficial for the range of sampling to change in more or less the same way during a set of updates, even when the number of updates vary. The constant 1000 here can also be set differently, but that will change the the best $\eta$ values. With this formulation, we always do uniform sampling in the first update, and we always have $\eta^{K\frac{1000}{K}} = \eta^{1000}$ in the last update.

The effect of such a sampling formulation is twofold. The first effect is the first mini-batch will be uniformly sampled from the whole buffer, the second mini-batch will be uniformly sampled from the the whole buffer excluding a few of the oldest data points in the buffer, and as $k$ grows more of the older data gets excluded. Clearly, the more recent a data point is, the more likely it will get sampled. The second effect is that we are doing this in an ordered way: we first sample from all the data in the buffer, and gradually shrink the range of sampling to only sample from the most recent data. This scheme reduces the chance of over-writing  parameter changes made by new data with parameter changes made by old data. We hypothesize that this process will allow us to better approximate the value functions near recently-visited states, while still maintaining an acceptable approximation near states visited in the more distant past. 


Different $\eta$ values are desirable depending on how fast the agent is learning and how fast the past experiences become obsolete. When the agent is learning fast we want $\eta$ to be lower so that we put more emphasis on the newer data. When the agent is learning slowly, we want $\eta$ to be higher so that it becomes closer to uniform sampling and the agent can make use of more data points in the past. 
A simple solution is to anneal $\eta$ during training. Let $T$ be the total number of timesteps in training. Let $\eta_0$ and $\eta_T$ be the initial and final $\eta$ value. We can set $\eta_T = 1$ so that it anneals to uniform sampling. The $\eta$ we use for timestep $t$ is $\eta_t = \eta_0 + (\eta_T - \eta_0) \cdot \frac{t}{T}$.

\begin{figure}[t]
\centering
        \begin{subfigure}[b]{0.325\textwidth}
                \includegraphics[width=\linewidth]{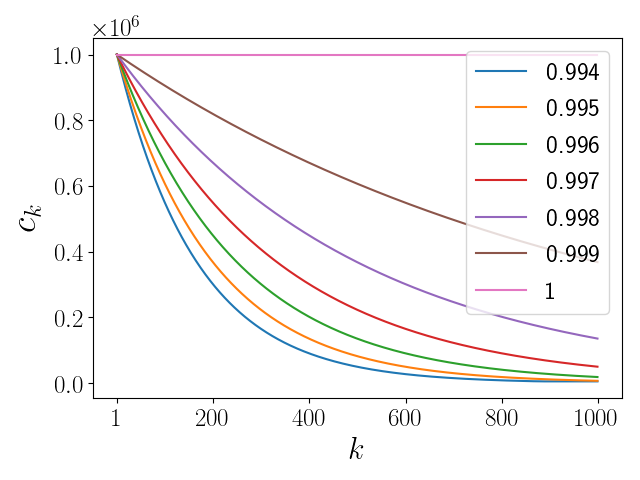}
                \caption{}
                \label{fig:eta_effect}
        \end{subfigure}%
        \begin{subfigure}[b]{0.325\textwidth}
                \includegraphics[width=\linewidth]{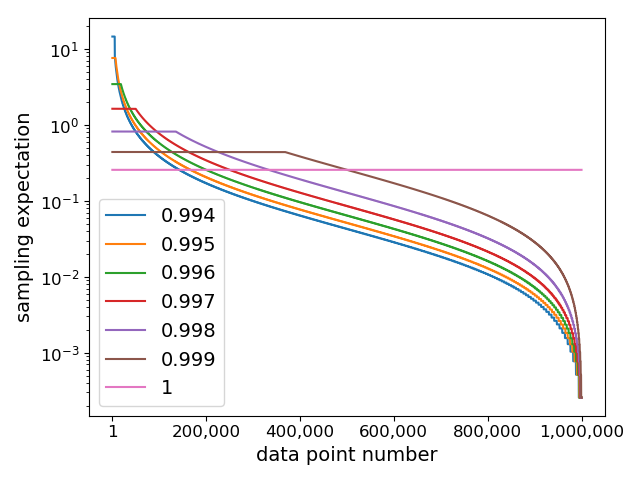}
                \caption{}
                \label{fig:data_expect}
        \end{subfigure}%
        \begin{subfigure}[b]{0.325\textwidth}
                \includegraphics[width=\linewidth]{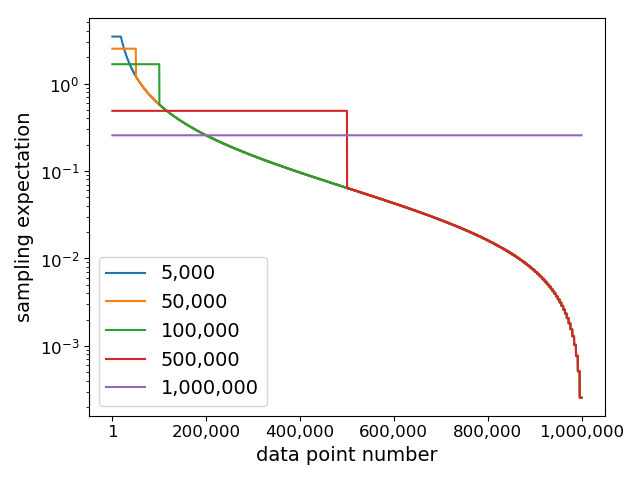}
                \caption{}
                \label{fig:data_expect_cmin}
        \end{subfigure}%
        \caption{Effect of different $\eta$ and $c_{min}$ values.
        The plots assume a replay buffer with 1 million samples, and 1,000 mini-batches in an update phase. Figure \ref{fig:eta_effect} plots  $c_k$ (ranging from 0 to 1 million) as a function of $k$ (ranging from 1 to 1,000). Figure \ref{fig:data_expect} and Figure \ref{fig:data_expect_cmin} plot the expected number of times a data point in the replay buffer is sampled for each data point in the replay buffer, with the data points ordered from most to least recent. Figure \ref{fig:data_expect} shows the expectations for different values of $\eta$, assuming 1000 updates each with a mini-batch size of 256, and $c_{min}=5000$. Figure \ref{fig:data_expect_cmin} shows the expectations for different values of $c_{min}$, assuming $\eta=0.996$. }
        \label{fig:eta_cmin_effect}
\end{figure}

Figure \ref{fig:eta_cmin_effect} shows how the $\eta$ and $c_{min}$ values affect the data sampling process. 
Figure \ref{fig:eta_effect} shows that within an update phase, the sampling range shrinks for each new mini-batch. 
In general, we found that $(0.994,0.999)$ is a good range for $\eta$. 
Figure \ref{fig:data_expect} shows that the expected number of samples of a given data point decreases from most recent to least recent data points. 
When $\eta=0.996$, the most recent data point has a sampling expectation that is about 10,000 times higher than the oldest data in buffer. Figure \ref{fig:data_expect_cmin} shows that when $\eta=0.996$, a large $c_{min}$ value increases the expected number of times an older data point is sampled. When $c_{min}$ equals the buffer size, we again obtain uniform sampling. 

\subsection{Soft Actor-Critic with Prioritized Experience Replay}
We also implement the proportional variant of Prioritized Experience Replay \cite{schaul2015prioritized} in SAC. 
Since SAC has two Q-networks, we redefine the absolute TD error $|\delta|$ of a transition $(s,a,r,s')$ to be the average absolute TD error of two Q networks:
\begin{equation}
|\delta| = \frac{1}{2} \sum_{l=1}^{2} |r + \gamma V_{\psi_{targ}}(s') - Q_{\theta,l}(s, a)|
\end{equation}
Within the sum, the first two terms $r + \gamma V_{\psi_{targ}}(s')$ is simply the target for the Q network, and the third term $Q_{\theta,l}(s, a)$ is the current estimate of the $l^{th}$ Q network. 
For the $i^{th}$ data point, the definition of the priority value $p_i$ is $p_i = |\delta_i| +\epsilon$.
The probability of sampling a data point $P(i)$ is computed as:
\begin{equation}
    P(i) = \frac{p_i^{\beta_1}}{\sum_j p_j^{\beta_1}}
\end{equation}
where $\beta_1$ is a hyperparameter that controls how much the priority value affects the sampling probability, which is denoted by $\alpha$ in \cite{schaul2015prioritized}, but to avoid confusion with the $\alpha$ in SAC, we denote it as $\beta_1$. The importance sampling (IS) weight $w_i$ for a data point is computed as:
\begin{equation}
    w_i=(\frac{1}{N} \cdot \frac{1}{P(i)})^{\beta_2}
\end{equation}
where $\beta_2$ is denoted as $\beta$ in \cite{schaul2015prioritized}. 

Based on the original SAC algorithm, we change the sampling method from uniform sampling to sampling using the probabilities $P(i)$, and for the Q updates we apply the IS weight $w_i$. This gives SAC with Prioritized Experience Replay (SAC+PER). 
We note that as compared with SAC+PER,  ERE does not require a special data structure and has negligible extra cost, while PER uses a sum-tree structure with some additional computational cost.  
We also tried several variants of SAC+PER, but preliminary results show that it is unclear whether there is improvement in performance, so we kept the algorithm simple.

\subsection{Soft Actor-Critic with Emphasizing Recent Experience and Prioritized Experience Replay}
We also propose a method that combines the above 2 methods (SAC+ERE+PER). This method does two things: first, during a set of mini-batch updates, the sampling range gradually shrinks as before. And second, from this sampling range, instead of uniformly sampling, we do priority sampling, where the sampling probability is proportional to the absolute TD-error of a data point. 

Assume we make $K$ mini-batch updates after some amount of data collection. Let $N$ be the max size of the replay buffer. Define $c_k$ as before. 
Let $D_{c_k}$ be the $c_k$ most recent data points in the replay buffer. Then the probability of sampling a data point is computed as:
\begin{equation}
    P(i) = \frac{p_i^\alpha}{\sum_j p_j^\alpha}, i, j\in D_{c_k}
\end{equation}
The priority value and importance sampling weight computation are the same as in SAC+PER.

\section{Mujoco experiments}

We perform experiments on a set of Mujoco \cite{todorov2012mujoco} environments implemented in OpenAI Gym \cite{brockman2016openai}. We aim to show how different experience replay schemes can affect the performance of SAC.
We are mainly concerned with four variants of SAC: vanilla SAC, SAC+ERE, SAC+PER and SAC+ERE+PER. For SAC+ERE, we pay special attention to how it affects the learning speed especially in early-stage. We perform additional experiments to show that the update order is important for SAC+ERE, and show how different hyperparameters can affect the performance of the SAC variants. 

To make our comparisons fair, analysis meaningful and results reproducible \cite{duan2016benchmarking,henderson2018matters, islam2017reproducibility}, for each variant we use the same SAC code base that we implemented in PyTorch, mainly based on the minimal SAC implementation in \cite{OpenAIspinup}. We use the same neural net architecture, activation function, optimizer, replay buffer size, learning rate and other hyper-parameters as reported in the SAC paper \cite{haarnoja2018soft} for the SAC baseline as well as for our three proposed enhancements to SAC. Note in the original SAC, all environments except Humanoid use the same reward scale. All other hyper-parameters are the same across environments. We run each set of experiments on ten random seeds. We run five evaluation episodes every 5000 data points. During evaluation episodes, we run the SAC policy deterministically, instead of sampling from the action distribution. For the plots, a solid line indicates the mean across 10 random seeds and the shaded area shows min and max values. Each point on the plot is smoothed over 50 evaluation episodes to make the figures easier to read. Additional implementation details can be found in the appendix. We will also post all code and data files online after proper cleaning and documentation. 

\subsection{SAC with smaller buffer size}
\begin{figure}[t]
\centering
        \begin{subfigure}[b]{0.325\textwidth}
                \includegraphics[width=\linewidth]{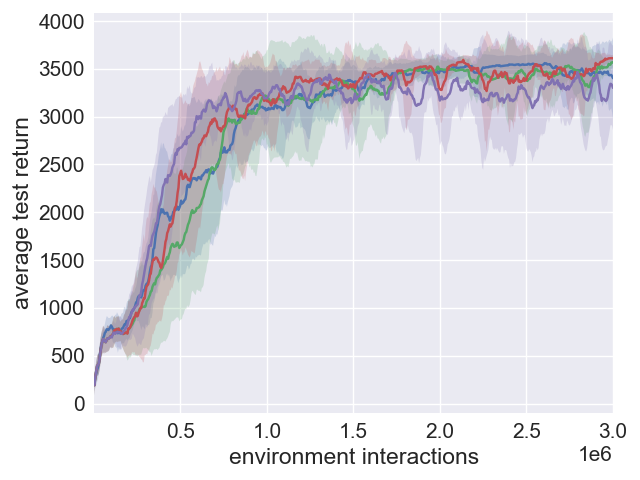}
                \caption{Hopper-v2}
                \label{fig:sac-sb-1}
        \end{subfigure}%
        \begin{subfigure}[b]{0.325\textwidth}
                \includegraphics[width=\linewidth]{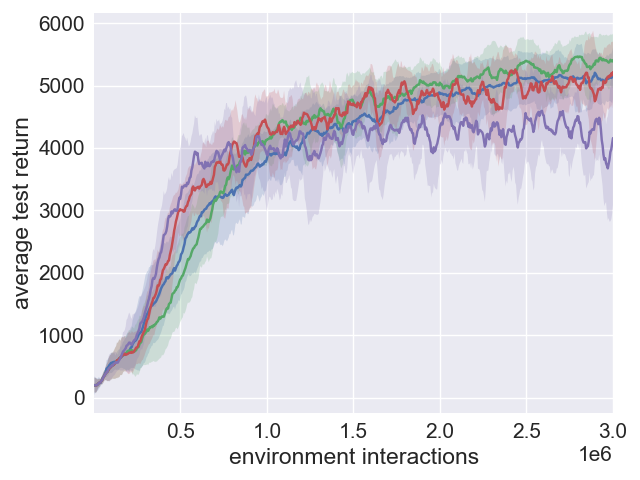}
                \caption{Walker2d-v2}
                \label{fig:sac-sb-2}
        \end{subfigure}%
        \begin{subfigure}[b]{0.325\textwidth}
                \includegraphics[width=\linewidth]{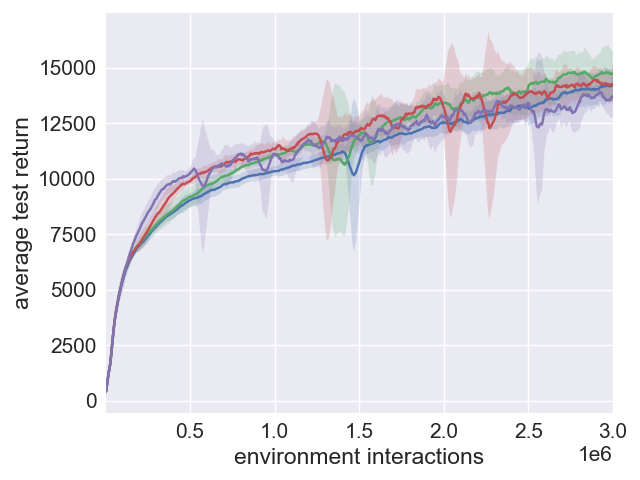}
                \caption{HalfCheetah-v2}
                \label{fig:sac-sb-3}
        \end{subfigure}\\
        \begin{subfigure}[b]{0.325\textwidth}
                \includegraphics[width=\linewidth]{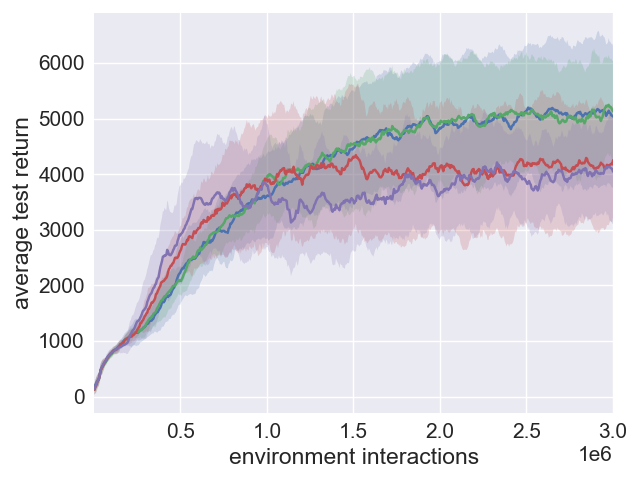}
                \caption{Ant-v2}
                \label{fig:sac-sb-4}
        \end{subfigure}
        \begin{subfigure}[b]{0.325\textwidth}
                \includegraphics[width=\linewidth]{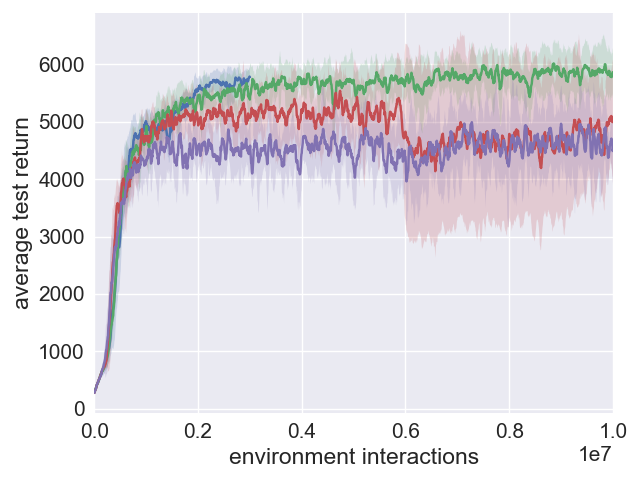}
                \caption{Humanoid-v2}
                \label{fig:sac-sb-5}
        \end{subfigure}
        \begin{subfigure}[b]{0.325\textwidth}
                \includegraphics[width=\linewidth]{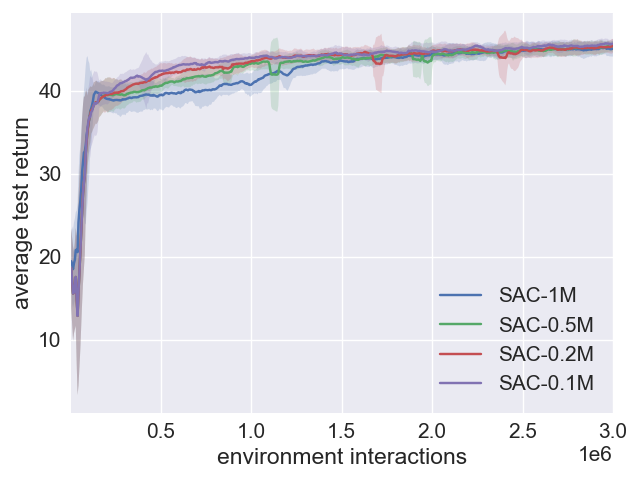}
                \caption{Swimmer-v2}
                \label{fig:sac-sb-6}
        \end{subfigure}
        
        \caption{SAC performance with different buffer sizes. In general, a smaller buffer can improve learning speed in the early stage, but will negatively affect performance in the long run. }
        \label{fig:sac-sb}
\end{figure}

As a motivating example, we first provide a set of experiments on SAC where the only difference is the buffer size. Figure \ref{fig:sac-sb} shows how different buffer sizes can affect the performance of the original SAC algorithm. We tested buffer sizes of 1M (baseline), 0.5M, 0.2M and 0.1M. Results show that a smaller buffer in general can make learning faster in the early stage, but at the same time can reduce the late-stage performance of the algorithm. For instance, in Ant-v2 and Walker2d-v2, a buffer size of 0.1M leads to the fastest learning in the first 0.75M data points, but then its performance plateaus and other variants with larger buffer size perform better. 

We hypothesize that a potential problem of using a small buffer is: since we only have a small amount of data, the neural networks in SAC might forget about how to perform the task well on some states visited earlier. This is a problem similar to catastrophic forgetting \cite{french1999catastrophic,mcclelland1995there, mccloskey1989catastrophic, ratcliff1990connectionist, robins1995catastrophic}, a term often used to refer to the situation where the agent has to learn two tasks A and B in a sequential order, and the knowledge about task A quickly gets forgotten as the agent starts to train on task B. We believe this issue also arises in the case of an RL agent learning a single highly-complex task. When using a small buffer, the agent might be able to learn well how to act in states that are stored in the buffer, but forget about the states that have been removed from the buffer. 


\subsection{SAC with Emphasizing Recent Experience}
\begin{figure}[t]
\centering
        \begin{subfigure}[b]{0.325\textwidth}
                \includegraphics[width=\linewidth]{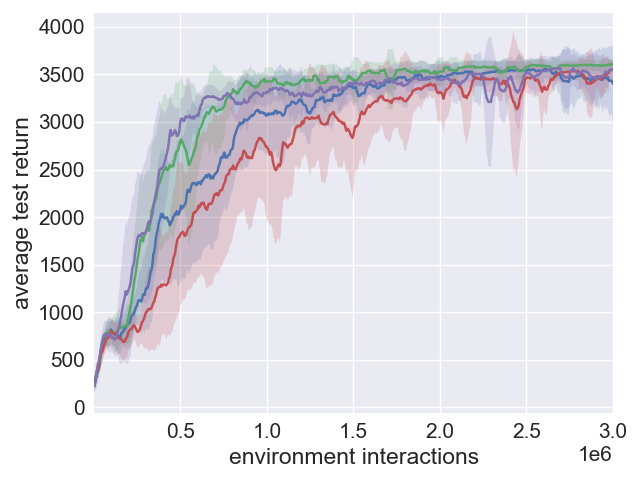}
                \caption{Hopper-v2}
                \label{fig:sac4-1}
        \end{subfigure}%
        \begin{subfigure}[b]{0.325\textwidth}
                \includegraphics[width=\linewidth]{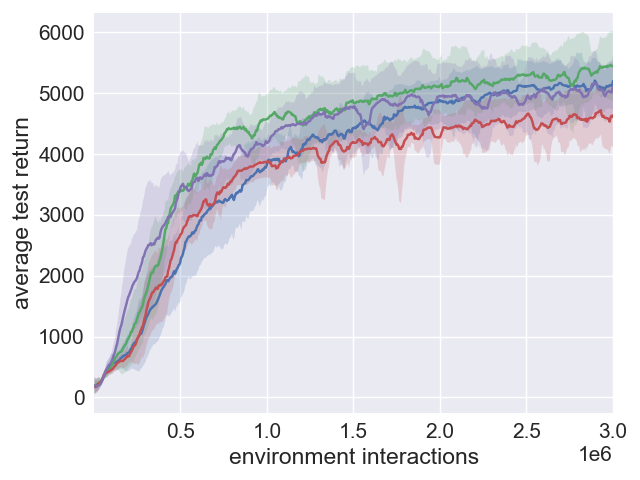}
                \caption{Walker2d-v2}
                \label{fig:sac4-2}
        \end{subfigure}%
        \begin{subfigure}[b]{0.325\textwidth}
                \includegraphics[width=\linewidth]{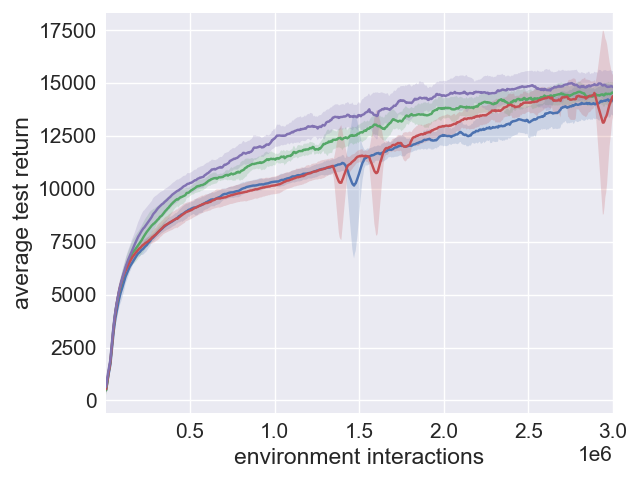}
                \caption{HalfCheetah-v2}
                \label{fig:sac4-3}
        \end{subfigure}\\
        \begin{subfigure}[b]{0.325\textwidth}
                \includegraphics[width=\linewidth]{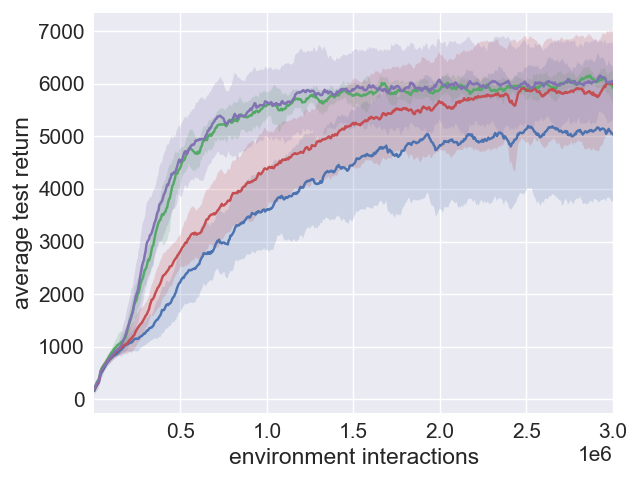}
                \caption{Ant-v2}
                \label{fig:sac4-4}
        \end{subfigure}
        \begin{subfigure}[b]{0.325\textwidth}
                \includegraphics[width=\linewidth]{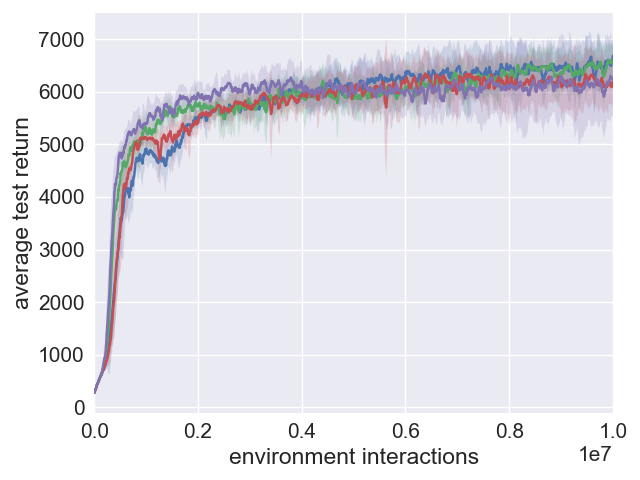}
                \caption{Humanoid-v2}
                \label{fig:sac4-5}
        \end{subfigure}
        \begin{subfigure}[b]{0.325\textwidth}
                \includegraphics[width=\linewidth]{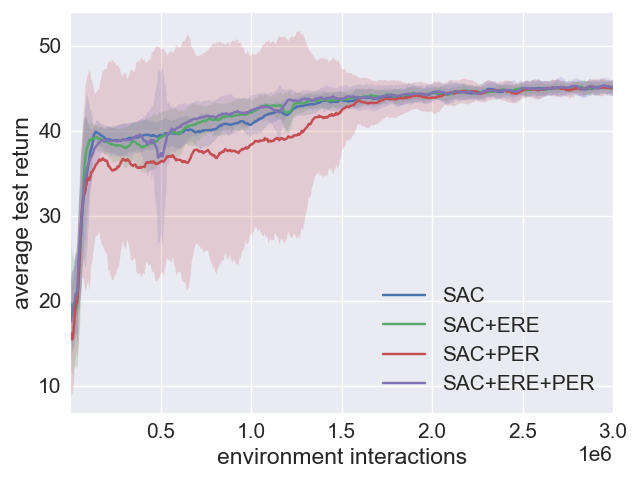}
                \caption{Swimmer-v2}
                \label{fig:sac4-6}
        \end{subfigure}
        \caption{Performance comparison of SAC, SAC+ERE, SAC+PER and SAC+ERE+PER}\label{fig:sac4}
\end{figure}

Figure \ref{fig:sac4} shows the performance of the variants of SAC on 6 different Mujoco environments. 
We first focus our analysis on the performance of SAC+ERE (green) compared with the SAC baseline (blue). 

For SAC+ERE we chose $\eta=0.996$ for all environments. The hyperparameter $\eta$ is obtained through preliminary hyperparameter search on Ant-v2. For all other hyperparameters, we use exactly those in the original SAC paper \cite{haarnoja2018soft}. The result shows that SAC+ERE consistently outperforms the SAC baseline in all environments and in all stages of training. For instance, in Ant-v2, SAC+ERE is 3 times faster to reach an average performance of 4500 compared to SAC, and it reaches 5500 at one million samples, while vanilla SAC never reaches 5500 in the first three million samples. In Hopper-v2, SAC+ERE is 1.5 times faster to reach 2500 compared to SAC. In Walker2d-v2, SAC+ERE is 1.5 times faster to reach 3000, in HalfCheetah-v2, SAC+ERE is 1.5 times faster to reach 10,000. Note that for SAC+ERE, we anneal $\eta$ to 1 linearly, which gives uniform sampling in the end; we therefore expect its performance to be the same as SAC when trained sufficiently long. 

\begin{figure}[t]
\centering
        \begin{subfigure}[b]{0.325\textwidth}
                \includegraphics[width=\linewidth]{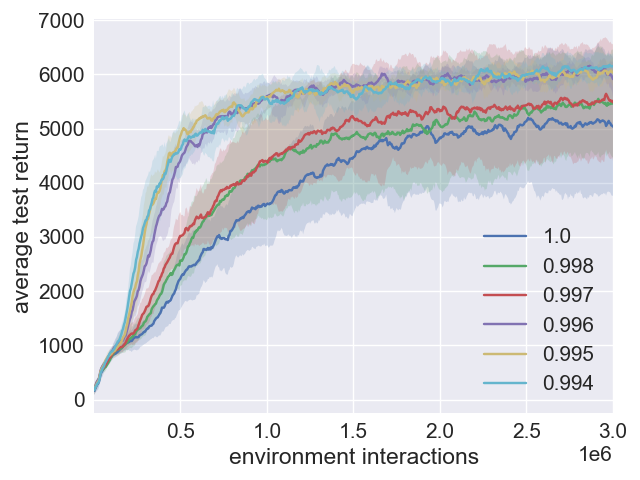}
                \caption{SAC+ERE $\eta$ values}
                \label{fig:sac-ere-hyp-sub}
        \end{subfigure}%
        \begin{subfigure}[b]{0.325\textwidth}
                \includegraphics[width=\linewidth]{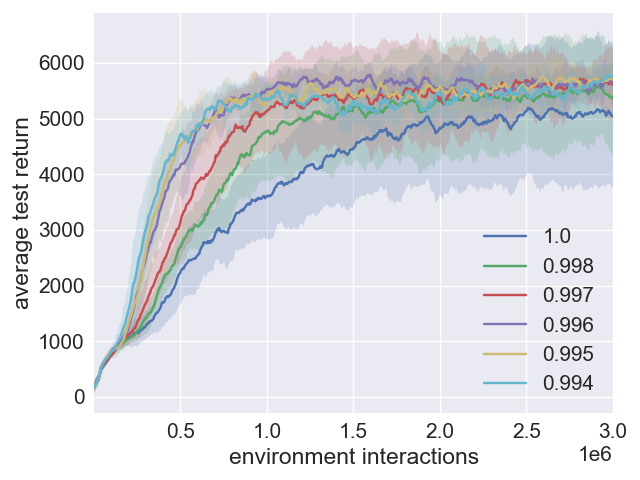}
                \caption{SAC+ERE no anneal}
                \label{fig:sac-ere-anneal}
        \end{subfigure}
        \begin{subfigure}[b]{0.325\textwidth}
                \includegraphics[width=\linewidth]{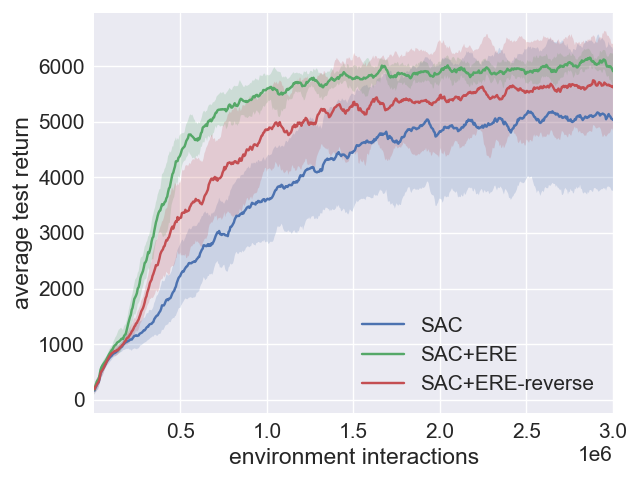}
                \caption{SAC+ERE update order}
                \label{fig:sac-ere-update-order}
        \end{subfigure}%
        \caption{Performance of SAC+ERE with different $\eta$ values, without annealing and with different update order, on Ant-v2.}\label{fig:sac-ere-hyp}
\end{figure}

We also found that SAC+ERE is relatively robust to the hyperparameter $\eta$. We found that any $\eta$ value in the range of $(0.994,0.999)$ consistently improves performance on all Mujoco environments, and especially in the early stages. Figure \ref{fig:sac-ere-hyp-sub} shows how different $\eta$ values can affect performance of SAC+ERE on Ant-v2. When using a large $\eta$ value it is similar to uniform sampling, so the learning becomes slower; and a small value such as $0.994$ can lead to very fast learning in the beginning. 

Figure \ref{fig:sac-ere-anneal} shows that annealing $\eta$ can improve robustness and long term performance of the SAC+ERE. Note that compared with results in Figure \ref{fig:sac-ere-hyp-sub}, which has annealing $\eta$, not annealing $\eta$ makes early stage learning even faster, but gives worse result in the long run. For instance, when $\eta=0.996$, SAC+ERE with annealing can reach an average score of 6000 near 3M, while without annealing it fluctuates around 5500. 

Our results also show that update order is indeed critical to improved performance. Figure \ref{fig:sac-ere-update-order} shows how different update orders can affect the performance of SAC+ERE on Ant-v2. We can see that SAC+ERE significantly outperform SAC in all stages of training. But if we reverse the update order, although the performance is still better than SAC, the average performance is greatly reduced in all stages of training compared to with the correct order. 
This shows that the two key components of ERE are both important to boost performance.

SAC is well-known to have excellent robustness properties \cite{haarnoja2018soft}, that is, the sample efficiency performance is not highly dependent on the initial seeds. 
Table \ref{tab:performance_robust_firsthalf} compares the robustness of SAC with SAC+ERE (as well as with other algorithms soon to be discussed).  At 1.5 million samples, we see that SAC+ERE has lower standard deviation than vanilla SAC for four of the six environments. Similar robustness metrics are considered in the appendix. We can conclude that ERE boosts the sample efficiency of SAC without compromising its robustness. 

\begin{table}[H]
\renewcommand{\arraystretch}{1.1}
\centering
\caption{SAC variants average performance and average std across seeds on the first 1.5M timesteps, highest performances for each environment are highlighted in boldface} 
\label{tab:performance_robust_firsthalf}
\vspace{1mm}
\begin{tabular}{ l l l l l }
\toprule
Environment 		&SAC 	&SAC+ERE & SAC+PER &SAC+ERE+PER\\
\midrule
Hopper-v2 	    &$2368.7 \pm554.6$ &$\mathbf{2705.5} \pm420.9$ &$2075.0 \pm645.8$ &$\mathbf{2731.0} \pm397.3$ \\
Walker2d-v2     &$2827.6 \pm632.5$ &$\mathbf{3444.0} \pm588.5$ &$2871.4 \pm437.5$ &$\mathbf{3413.6} \pm634.9$ \\
HalfCheetah-v2 	&$9113.8 \pm482.6$ &$10043.5 \pm582.8$ &$9116.0 \pm493.6$ &$\mathbf{10681.5} \pm659.7$ \\
Ant-v2 			   &$2775.1 \pm701.0$ &$\mathbf{4327.2} \pm515.9$ &$3326.0 \pm614.8$ &$\mathbf{4429.7} \pm754.3$ \\
Humanoid-v2 &$3511.1 \pm637.9$ &$4076.2 \pm606.2$ &$3693.5 \pm503.0$ &$\mathbf{4319.8} \pm507.4$ \\
Swimmer-v2       &$39.9 \pm2.5$ &$40.1 \pm2.8$ &$37.1 \pm11.4$ &$40.3 \pm3.2$ \\
\bottomrule
\end{tabular}
\end{table}



\subsection{SAC with Emphasizing Recent Experience and Prioritized Experience Replay}
We now analyze the performance of the other two SAC variants. For the hyperparameters for SAC+PER, we chose $\beta_1 = 0.6$, $\beta_2 = 0.6$, obtained through preliminary hyperparameter search on Ant-v2. We found that although a wide range of $\beta_1$ and $\beta_2$ values give performance gain on Ant-v2, they did not work on all environments. A more detailed analysis on hyperparameters for SAC+PER is given in the appendix.

From the results in Figure \ref{fig:sac4} we see that SAC+PER (red) significantly outperforms SAC (blue) on Ant-v2, which is the environment used to do hyperparameter search, and does better than SAC near the end of training on HalfCheetah-v2, but it does similar or worse compared to SAC in other environments. It seems that a good hyperparameter combination for SAC+PER can be very different across environments. 

SAC+ERE+PER (purple) can further boost early stage learning speed beyond the SAC+ERE boost, and sometimes can boost overall performance too. For instance, SAC+ERE+PER outperforms all other SAC variants on HalfCheetah-v2 in all stages of training, but it does similar to SAC+ERE, or somewhere in-between SAC+ERE and SAC+PER in other environments. 

\section{Conclusion}
We proposed Emphasizing Recent Experience, a new experience replay method that is simple but powerful. We showed it can significantly boost the learning speed of SAC, and in some environments it can also achieve better results in the long run. 
ERE is a general method that in theory can be applied to any off-policy DRL algorithm with a replay buffer.

We compared SAC+ERE with the popular Prioritized Experience Replay method and showed that ERE is easier to implement and does not require special data structures. With ERE the additional computation cost is negligible, and there is only one important hyperparameter, which we found to be easy to tune since a good hyperparameter found for one environment ($\eta \in (0.994,0.999)$) also works well in all environments.  We also showed that empirically in Mujoco environments, SAC+ERE has stronger performance than SAC+PER. However, it is possible that a more sophisticated formulation of SAC+PER can give better results.
We believe the two methods each have their unique strengths; for example, when the reward is sparse, we expect PER to do well, since PER by design is strong at tackling sparse reward situations while ERE focuses on emphasizing recent data. We then proposed SAC+ERE+PER, which is a combination of the ERE and PER, and showed that it achieves even better performance in some environments. However, this variant loses the simplicity of SAC+ERE and has some extra computation cost due to the PER part.

For future work, we plan to also test ERE on other off-policy DRL algorithms such as DQN and on other benchmarks such as the Atari games to see if the significant performance gains observed on Mujoco generalize to other algorithms and environments.


\bibliographystyle{plain}
\bibliography{reference.bib}


\appendix
\newpage
\section{Pseudocode}
In this section we give the pseudocode for the 3 SAC variants we proposed. For a minimal SAC pseudocode please check section 4.2 in \cite{haarnoja2018soft}, for PER please check section 3.3 in \cite{schaul2015prioritized}. Our pseudocode has been mainly based on the original SAC and PER pseudocode. We have modified some of the code structure to make it more similar to our actual implementation. And we give a large number of comments in our pseudocode to make sure each step is clear. 
Algorithm \ref{alg:sac-ere} shows the code for SAC+ERE, and algorithm \ref{alg:sac-ere-per} shows the code for SAC+ERE+PER. To obtain SAC+PER one can simply replace line 14, 15 in algorithm \ref{alg:sac-ere-per} with uniform sampling. For the computation of the loss functions and the gradients for networks in SAC, the following is a very short summary. Please refer to \cite{haarnoja2018soft} for theory and details. Note that in the pseudocode we use $\lambda$ to denote learning rate, we use the same learning rate for every network, and for the gradient update steps, we do not expand the gradient equations to make thing simpler. 

The loss for training the $V$ network is: 

\begin{align}
\label{eq:v_cost}
J_V(\vparams) = \E{\st \sim \mathcal{D}}{\frac{1}{2}\left(\V_\vparams(\st) - \E{\at\sim\policy_\pparams}{Q_\params(\st, \at) - \log \policy_\pparams(\at|\st)}\right)^2}
\end{align}

An unbiased estimator of the gradient of the above loss function is:

\begin{align}
\hat \nabla_\vparams J_V(\vparams) = \nabla_\vparams \V_\vparams(\st) \left(\V_\vparams(\st) - Q_\params(\st, \at) + \log \policy_\pparams(\at|\st)\right)
\label{eq:v_gradient}
\end{align}

The loss for training the $Q$ network is: 
\begin{align}
J_\Q(\params) = \E{(\st, \at)\sim\mathcal{D}}{\frac{1}{2}\left(\Q_\params(\st, \at) - \hat \Q(\st, \at)\right)^2},
\label{eq:q_cost}
\end{align}
with
\begin{align}
\hat \Q(\st, \at) = \reward(\st, \at) + \discount \E{\stp\sim\pdyn}{\V_\vtargetparams(\stp)},
\end{align}

The gradient of the above loss function can be computed as:

\begin{align}
\hat \nabla_\params J_Q(\params) =  \nabla_\params \Q_\params(\at, \st) \left(\Q_\params(\st, \at) - \reward(\st, \at) - \discount \V_\vtargetparams(\stp)\right)
\end{align}

Where the target value network  $\V_\vtargetparams$ is an exponentially moving average of the value network. 

The loss for the policy network is the expected KL-Divergence: 
\begin{align}
J_\policy(\pparams) = \E{\st\sim\mathcal{D}}{\kl{\policy_\pparams(\voidarg|\st)}{\frac{\exp\left(Q_\params(\st, \voidarg)\right)}{Z_\params(\st)}}}.
\label{eq:policy_objective}
\end{align}

After applying the reparameterization trick:

\begin{align}
\at = f_\pparams(\epsilon_t; \st),
\end{align}

The loss now becomes:

\begin{align}
J_\policy(\pparams) = \E{\st\sim\mathcal{D},\epsilon_t\sim\gauss}{\log \policy_\pparams(f_\pparams(\epsilon_t;\st)|\st) - Q_\params(\st, f_\pparams(\epsilon_t;\st))}
\label{eq:reparam_objective}
\end{align}

And the gradient can be computed as: 

\begin{align}
\hat\nabla_\pparams J_\policy(\pparams) = \nabla_\pparams \log \policy_\pparams(\at|\st) \notag + (\nabla_\at \log \policy_\pparams(\at|\st) - \nabla_\at Q(\st, \at))\nabla_\pparams f_\pparams(\epsilon_t;\st)
\label{eq:policy_gradient}
\end{align}

\begin{algorithm} 
\caption{Soft Actor Critic with Emphasizing Recent Experience}
\label{alg:sac-ere}
\begin{algorithmic}[1]
\State Initialize parameter vectors $\psi$, $\bar{\psi}$, $\theta$, $\phi$
\State Initialize timestep $t = 1$, episode length $K = 0$
\State Get initial state from environment $s_{1}  \sim p(s_{init})$
\For{$t = 1,\dots,T$} \Comment{T is 10M for humanoid, 3M for other environments}
\State $a_t \sim \pi_\phi(a_t | s_t)$  \Comment{sample action from policy}
\State $s_{t+1} \sim p(s_{t+1}|s_t, a_t)$ \Comment{sample transition from environment}
\State $D \leftarrow D \cup \{(s_t, a_t, r(s_t, a_t), s_{t+1}) \}$ \Comment{add transition to replay buffer}
\State $\eta_t = \eta_0 + (\eta_T - \eta_0) \cdot \frac{t}{T}$ \Comment{compute the annealed $\eta_t$}
\State $t \leftarrow t + 1$ 
\State $K \leftarrow K + 1$ 
\If{$s_{t+1}$ is a terminal state} \Comment{number of updates is the same as episode length $K$}
\For{$k= 1,\dots,K$ mini-batch update}
\State $c_k = N \cdot \eta_t^{k\frac{1000}{K}}$ \Comment{compute the sampling range}
\State $B \sim D_{c_k}$ \Comment{sample a mini-batch uniformly from buffer's sampling range}
\State The following network updates are computed on the mini-batch $B$
\State $\psi \leftarrow \psi - \lambda \hat{\nabla}_\psi J_V(\psi)$ \Comment{update value network}
\State $\theta_l \leftarrow \theta_l - \lambda \hat{\nabla}_{\theta_l} J_Q(\theta_l)$ for $l\in \{1,2\}$ \Comment{update Q networks}
\State $\phi \leftarrow \phi - \lambda \hat{\nabla}_\phi J_\pi(\phi)$ \Comment{update policy network}
\State $\bar{\psi} \leftarrow \tau \psi + (1-\tau) \bar{\psi}$ \Comment{update target value network}
\EndFor 
\State $K = 0$
\State $s_{t+1}  \sim p(s_{init})$ \Comment{reset environment, get new initial state}
\EndIf
\EndFor
\end{algorithmic}
\end{algorithm}

\begin{algorithm} 
\caption{Soft Actor Critic with Emphasizing Recent Experience and Prioritized Experience Replay}
\label{alg:sac-ere-per}
\begin{algorithmic}[1]
\State Initialize parameter vectors $\psi$, $\bar{\psi}$, $\theta$, $\phi$
\State Initialize timestep $t = 1$, episode length $K = 0$
\State Get initial state from environment $s_{1}  \sim p(s_{init})$
\For{$t = 1,\dots,T$} \Comment{T is 10M for humanoid, 3M for other environments}
\State $a_t \sim \pi_\phi(a_t | s_t)$  \Comment{sample action from policy}
\State $s_{t+1} \sim p(s_{t+1}|s_t, a_t)$ \Comment{sample transition from environment}
\State $D \leftarrow D \cup \{(s_t, a_t, r(s_t, a_t), s_{t+1}) \}$ \Comment{add transition to replay buffer}
\State $\eta_t = \eta_0 + (\eta_T - \eta_0) \cdot \frac{t}{T}$ \Comment{compute the annealed $\eta_t$}
\State $t \leftarrow t + 1$ 
\State $K \leftarrow K + 1$ 
\If{$s_{t+1}$ is a terminal state} \Comment{number of updates is the same as episode length $K$}
\For{$k= 1,\dots,K$ mini-batch update}
\State Set $\Delta \psi, \Delta \theta, \Delta \phi =\mathbf{0}$
\State $c_k = N \cdot \eta_t^{k\frac{1000}{K}}$ \Comment{compute the sampling range}
\State $B \sim D_{c_k}$, $P(i) = \frac{p_i^\alpha}{\sum_j p_j^\alpha}$, $j \in D_{c_k}$  \Comment{sample a mini-batch with probabilities $P(i)$}
\For{$b \in B$} \Comment{for each data point in mini-batch $B$}
\State $w_b=(\frac{1}{N} \cdot \frac{1}{P(b)})^{\beta_2}/{max_j w_j}$ \Comment{compute importance sampling weight}
\State $|\delta_b| = \frac{1}{2} \sum_{l=1}^{2} |r + \gamma V_{\psi_{targ}}(s_{b+1}) - Q_{\phi,l}(s_b, a_b)|$ \Comment{get abs TD error}
\State $p_b \leftarrow |\delta_b| +\epsilon$ \Comment{update priority}
\State The following 3 lines accumulate weight-change computed on transition $b$.
\State $\Delta\psi \leftarrow \Delta\psi + w_b  \hat{\nabla}_\psi J_V(\psi)$ 
\State $\Delta\theta_l \leftarrow \Delta\theta_l + w_b  \hat{\nabla}_{\theta_l} J_Q(\theta_l)$, for $l\in \{1,2\}$
\State $\Delta\phi \leftarrow \Delta\phi +  w_b \hat{\nabla}_\phi J_\pi(\phi)$ 
\EndFor

\State $\psi \leftarrow \psi - \lambda \Delta\psi$ \Comment{update value network}
\State $\theta_l \leftarrow \theta_l - \lambda \Delta\theta_l $, for $l\in \{1,2\}$ \Comment{update Q networks}
\State $\phi \leftarrow \phi -\lambda \Delta\phi$ \Comment{update policy network}
\State $\bar{\psi} \leftarrow \tau \psi + (1-\tau) \bar{\psi}$ \Comment{update target value network}
\EndFor 
\State $K = 0$
\State $s_{t+1}  \sim p(s_{init})$ \Comment{reset environment, get new initial state}
\EndIf
\EndFor
\end{algorithmic}
\end{algorithm}

\section{Implementation details and hyperparameters}
\label{app:hypers}
Here we give implementation details and list the hyperparameters we used to run the experiments. 

\subsection{Details in SAC implementation}
We first give a list of details in our codebase to facilitate reproduction of code. We implemented SAC using PyTorch, and our code structure has mainly followed the clear explanation on \cite{OpenAIspinup}. We used reparameterization trick to generate the action from the policy, for the log probability computation of the actions, we used the technique described in enforcing action bounds section in the SAC paper. Since our policy network gives action in the range $(-1, 1)$, we obtain from each environment an action limit value (how big the magnitude of the action can be) and when our network outputs an action in range $(-1, 1)$, the action is multiplied with the action limit to give an action that is in the action range of the environment. 

One important difference from the original SAC code base is, in original SAC, a data collection step is immediately followed by a mini-batch update. While in our case we first collect an episode of data points (could be anywhere between 1 to 1000 data points), and then we do a number of mini-batch updates, the number of mini-batch is the same as the number of data points collected in that episode. Compared to original SAC, throughout training we collect the same number of data points, and take the same total number of mini-batch updates, for example, in Ant-v2, this is 3M data points and 3M updates. It's unclear which way of doing the updates is more beneficial, we found that our SAC baseline to be slightly stronger than original SAC in some environments and slightly weaker in others, the difference is very small. However, enforcing an update order makes more sense when we do a set of updates while the agent is not interacting with the environment. It might be possible to formulate a novel ordered update scheme in the case of one data, one update, but this will be left as future work. 

\subsection{SAC hyperparameters}
All hyperparameters related to original SAC are same as used in the original SAC paper. For the $\alpha$ value, we set it to be 0.05 for Humanoid-v2 and 0.2 for all other OpenAI Mujoco environments, as given in the original SAC paper. This is given in table \ref{tab:shared_params}.

\begin{table}[tb]
\renewcommand{\arraystretch}{1.1}
\centering
\caption{SAC hyperparameters}
\label{tab:shared_params}
\vspace{1mm}
\begin{tabular}{l l| l }
\toprule
\multicolumn{2}{l|}{Parameter} &  Value\\
\midrule
\multicolumn{2}{l|}{\it{SAC}}& \\
& optimizer &Adam \cite{kingma2014adam}\\
& learning rate & $3 \cdot 10^{-4}$\\
& discount ($\discount$) &  0.99\\
& replay buffer size & $10^6$\\
& number of hidden layers (all networks) & 2\\
& number of hidden units per layer & 256\\
& number of samples per minibatch & 256\\
& nonlinearity & ReLU\\
& target smoothing coefficient ($\tau$)& 0.005\\
& target update interval & 1\\
\midrule
\multicolumn{2}{l|}{\it{ERE}}& \\
& $\eta_0$ & 0.996\\
& $\eta_T$ & 1.0\\
\midrule
\multicolumn{2}{l|}{\it{PER}}& \\
& $\beta_1$ & 0.6\\
& $\beta_2$ & 0.6\\
\bottomrule
\end{tabular}
\end{table}

\begin{table}[tb]
\renewcommand{\arraystretch}{1.1}
\centering
\caption{SAC environment specific parameters}
\label{tab:env_params}
\vspace{1mm}
\begin{tabular}{ l l l l l }
\toprule
Environment 		&Action Dimensions	&$\alpha$\\
\midrule
Hopper-v2 			&3		& 0.2\\
Walker2d-v2 		&6 		& 0.2\\
HalfCheetah-v2 		&6		& 0.2\\
Ant-v2 				&8		& 0.2\\
Swimmer-v2          &2      & 0.2\\
Humanoid-v2 	    &17 	& 0.05\\
\bottomrule
\end{tabular}
\end{table}

\subsection{Hyperparameters of SAC+ERE}
The hyperparameter choice for SAC+ERE is decided with some hyperparameter search on Ant-v2. We first reasoned that we should start searching the range $(0.990, 1.0)$, since smaller values of $\eta$ likely will put too much emphasis on the most recent data and breaks performance. We then found that a value in the range $(0.994, 0.999)$ give improvements on Ant-v2. And they also seem to work pretty well in other Mujoco environments as well. We did not fine tune SAC related hyperparameters for SAC+ERE, to showcase what performance gain we can obtain by simply changing the replay scheme to ERE. 

\subsection{Hyperparameters of SAC+PER}
\begin{figure}[tb]
\centering
        \begin{subfigure}[b]{0.325\textwidth}
                \includegraphics[width=\linewidth]{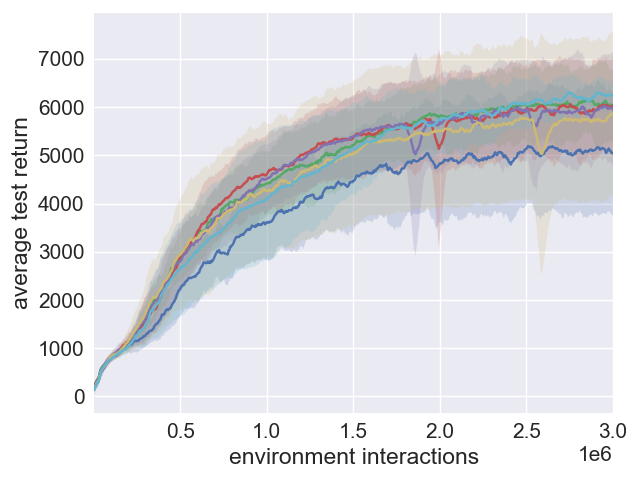}
                \caption{$\beta_2 = 0.4$}
                \label{fig:per-hyp-1}
        \end{subfigure}%
        \begin{subfigure}[b]{0.325\textwidth}
                \includegraphics[width=\linewidth]{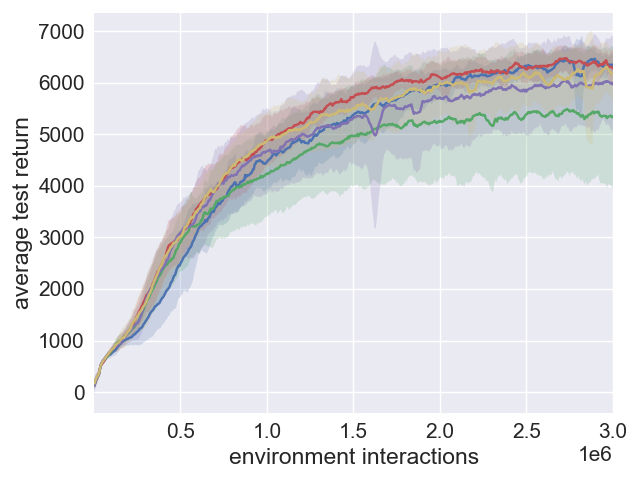}
                \caption{$\beta_2 = 0.5$}
                \label{fig:per-hyp-2}
        \end{subfigure}%
        \begin{subfigure}[b]{0.325\textwidth}
                \includegraphics[width=\linewidth]{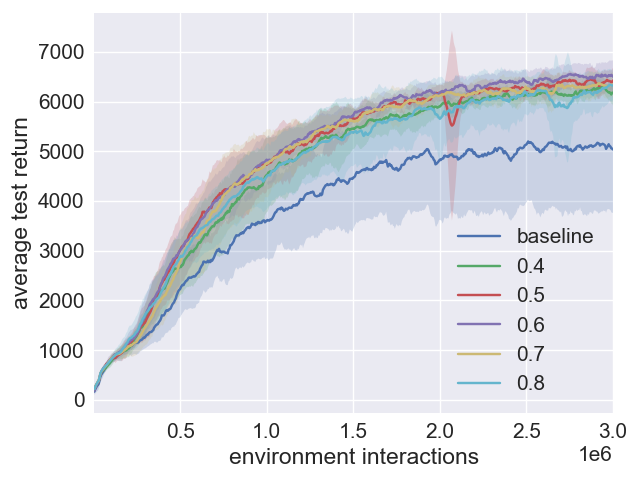}
                \caption{$\beta_2 = 0.6$}
                \label{fig:per-hyp-3}
        \end{subfigure}
        \caption{SAC+PER Performance with different $\beta_1$ and $\beta_2$ values on Ant-v2, the legend shows different $\beta_1$ values.}\label{fig:sac-per-hyp}
\end{figure}

Figure \ref{fig:sac-per-hyp} shows how different hyperparameter settings can affect training of SAC+PER in Ant-v2 environment. We mainly look at the $\beta_1$ and $\beta_2$ hyper-parameter. When we compared all the results together, the best setting on average is $\beta_1 = 0.6$ and $\beta_2 = 0.6$. So we use these values across all experiments. Note that some other hyperparameter settings give better performance on some seeds, but not better on average. Although these values work well in Ant-v2, they don't seem to work too well for the other environments.

\begin{figure}[tb]
\centering
        \begin{subfigure}[b]{0.325\textwidth}
                \includegraphics[width=\linewidth]{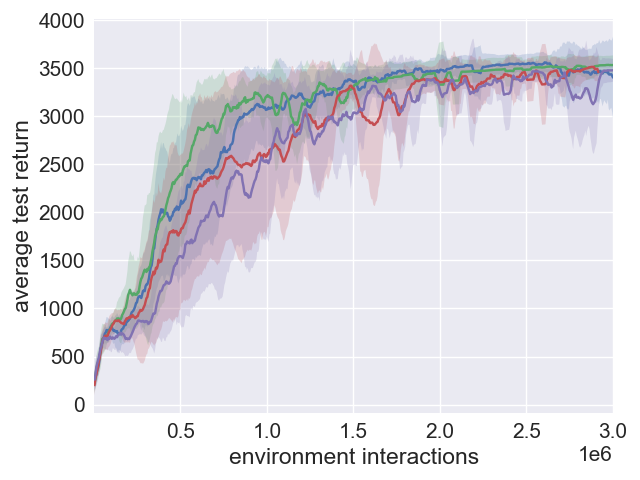}
                \caption{$\beta_2 = 0.2$}
                \label{fig:per-hyph-1}
        \end{subfigure}%
        \begin{subfigure}[b]{0.325\textwidth}
                \includegraphics[width=\linewidth]{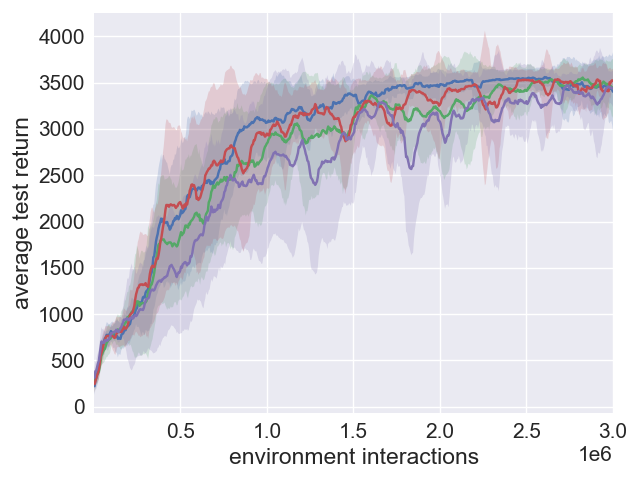}
                \caption{$\beta_2 = 0.4$}
                \label{fig:per-hyph-2}
        \end{subfigure}%
        \begin{subfigure}[b]{0.325\textwidth}
                \includegraphics[width=\linewidth]{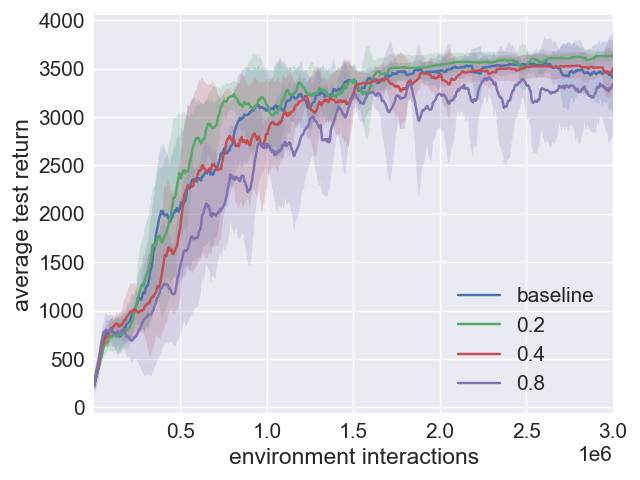}
                \caption{$\beta_2 = 0.8$}
                \label{fig:per-hyph-3}
        \end{subfigure}
        \caption{SAC+PER Performance with different $\beta_1$ and $\beta_2$ values on Hopper-v2, the legend shows different $\beta_1$ values.}\label{fig:sac-per-hyph}
\end{figure}

\begin{figure}[tb]
\centering
        \begin{subfigure}[b]{0.325\textwidth}
                \includegraphics[width=\linewidth]{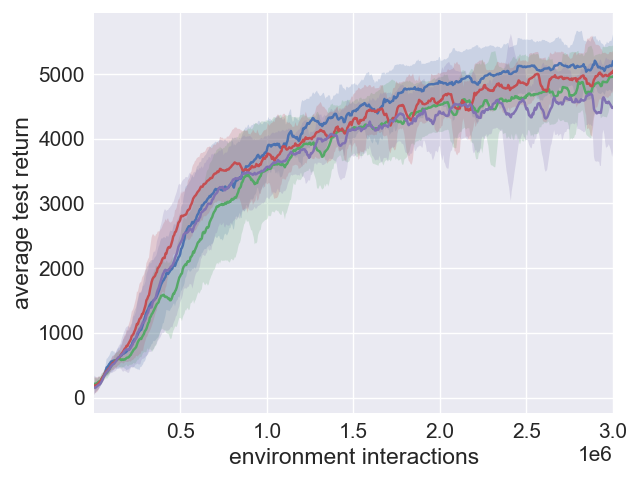}
                \caption{$\beta_2 = 0.2$}
                \label{fig:per-hypw-1}
        \end{subfigure}%
        \begin{subfigure}[b]{0.325\textwidth}
                \includegraphics[width=\linewidth]{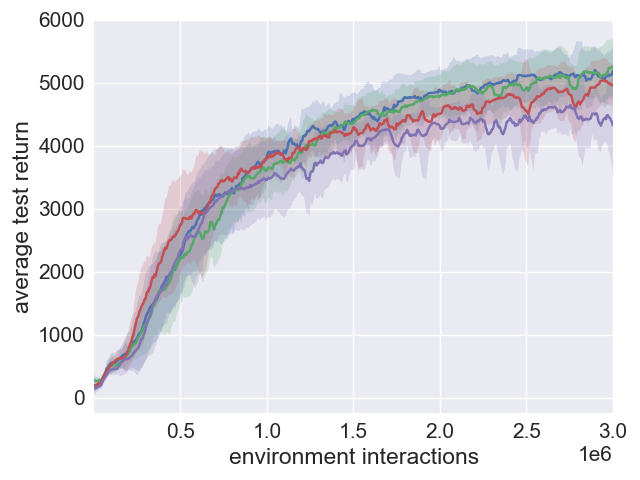}
                \caption{$\beta_2 = 0.4$}
                \label{fig:per-hypw-2}
        \end{subfigure}%
        \begin{subfigure}[b]{0.325\textwidth}
                \includegraphics[width=\linewidth]{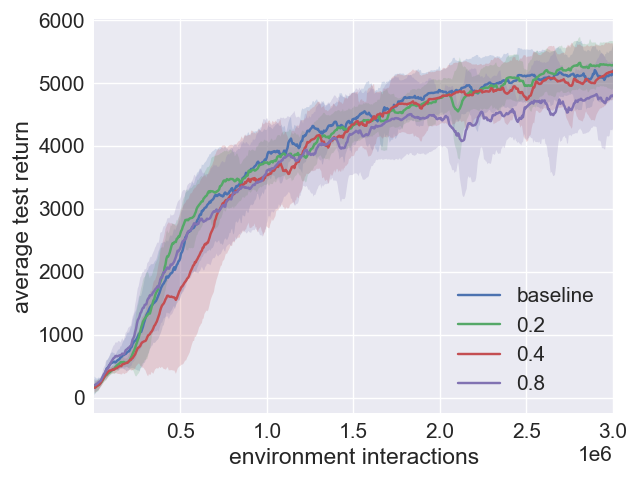}
                \caption{$\beta_2 = 0.8$}
                \label{fig:per-hypw-3}
        \end{subfigure}
        \caption{SAC+PER Performance with different $\beta_1$ and $\beta_2$ values on Walker2d-v2, the legend shows different $\beta_1$ values.}\label{fig:sac-per-hypw}
\end{figure}

Figure \ref{fig:sac-per-hyph} and \ref{fig:sac-per-hypw} show additional hyperparameter search on Hopper-v2 and Walker2d-v2. We found that it can be relatively difficult to find a good hyperparameter combination for SAC+PER. Fine tuning on each environment extensively can indeed improve performance, but the hyperparameter search is much more difficult compared to SAC+ERE. 

We have also tried reduce the learning rate to $1/4$ and $1/2$ of the original learning rate, since this was done in the PER paper \cite{schaul2015prioritized}, but our preliminary results show no significant improvement in performance. It's possible that we might need to perform a more extensive hyperparameter search for SAC+PER in order to get better results. 

\subsection{Hyperparameters of SAC+ERE+PER}
For the hybrid algorithm, we did not fine tune its hyperparameters, but used the same values from SAC+ERE and SAC+PER. 

\section{Robustness of SAC versus SAC+ERE}

Table \ref{tab:performance_robust} and table \ref{tab:performance_robust_new} further compare the robustness of SAC+ERE with the robustness of SAC. We again see that ERE does not compromise the robustness of SAC.

\begin{table}[H]
\renewcommand{\arraystretch}{1.1}
\centering
\caption{SAC variants average test episode return and average std across seeds, the average test episode return is a measure for performance, computed as the average test return across every seed and every evaluation episode. The average std across seeds is a measure for robustness over random seeds, is the std of test episode return, computed across every seed and then averaged over each evaluation episode. We run 5 evaluation episodes after 5000 environment interactions} 
\label{tab:performance_robust}
\vspace{1mm}
\begin{tabular}{ l l l l l }
\toprule
Environment 		&SAC 	&SAC+ERE & SAC+PER &SAC+ERE+PER\\
\midrule
Hopper 		&$2921.6 \pm383.3$ &$\mathbf{3131.5} \pm270.7$ &$2718.6 \pm482.2$ &$\mathbf{3091.7} \pm328.5$ \\
Walker2d 	&$3867.6 \pm576.3$ &$\mathbf{4305.2} \pm603.9$ &$3653.1 \pm466.0$ &$4160.5 \pm666.0$ \\
HalfCheetah 	&$10993.8 \pm625.1$ &$11970.0 \pm648.4$ &$11188.4 \pm697.9$ &$\mathbf{12579.5} \pm783.2$ \\
Ant 			    &$3856.2 \pm991.1$ &$\mathbf{5127.3} \pm519.2$ &$4498.8 \pm818.1$ &$\mathbf{5201.1} \pm825.7$ \\
Humanoid 	&$5722.5 \pm599.7$ &$5768.6 \pm585.9$ &$5647.8 \pm542.5$ &$5796.7 \pm636.9$ \\
Swimmer        &$42.2 \pm1.9$ &$42.4 \pm1.9$ &$40.7 \pm6.6$ &$42.4 \pm2.3$ \\
\bottomrule
\end{tabular}
\end{table}

\begin{table}[H]
\renewcommand{\arraystretch}{1.1}
\centering
\caption{Timestep and std across seeds for SAC and SAC+ERE when they reach a target score of 80\% SAC baseline final performance, baseline final performance is computed as SAC baseline's average test return in the last 0.1M timestep of training, std is computed for test return in nearby 0.1M timestep of reaching the target score.}
\label{tab:performance_robust_new}
\vspace{1mm}
\begin{tabular}{ l l l l l l l}
\toprule
Environment 		&Hopper 	&Walker2d  &HalfCheetah    &Ant  &Humanoid &Swimmer\\
\midrule
target score  &2759.6 &4087.7   &11321.3   &4095.9     &5283.6   &36.1\\  
SAC timestep          &770,000 &1,080,000  &1,420,000    &1,130,000    &1,510,000 &85,000\\
ERE timestep      &485,000  &615,000  &935,000     &450,000     &835,000 &70,000\\
SAC std              &717.8 &591.7     &1341.5      &978.1      &637.2  &9.0\\ 
ERE std          &829.7 &672.9    &525.4        &791.5      &701.1  &9.3\\
\bottomrule
\end{tabular}
\end{table}

\section{Computing infrastructure}
We mainly run our experiments on cpu nodes of a high-performance computer cluster, the specification of a single cpu node is: Intel(R) Xeon(R) CPU E5-2620 v3 @ 2.40GHz. Each job is run on a single cpu node until completion. 

\section{Programming and computation complexity}
In this section we give a more detailed analysis of the additional programming and computation complexity that are added to SAC by our proposed experience replay schemes.

In terms of programming complexity, SAC+ERE is a clear winner since it only requires a small adjustment to how your buffer sample mini-batches. It doesn't modify how the buffer store the data, and doesn't require special data structure to make it work efficiently. Thus the implementation difficulty is minimal. PER (proportional variant) requires a sum-tree data structure to make it run efficiently. The implementation is not too complicated, but compared to ERE it's a lot more work.

In terms of computation complexity (not sample efficiency), and wall-clock time, ERE's extra computation is negligible. For each mini-batch update we only need to compute one $c_k$ value, and annealing $\eta$ is also just a constant cost operation. In practice we observe no difference in computation time between SAC and SAC+ERE. On Ant-v2 with 3M data points, SAC takes 25-30 hours to run, and SAC-ERE takes about the same time. PER needs to update the priority of its data points constantly and compute sampling probability for all the data points. The complexity for sampling and updates is $O(log(N))$, the rank-based variant is similar \cite{schaul2015prioritized}. Although this is not too bad, it does impose a significant overhead on SAC, also note that this overhead grows linearly with the size of mini-batch. In our experiments, SAC+PER on Ant-v2 with 3M data can take up to 40 hours to run. SAC+ERE+PER runs with the same computation as SAC+PER.

\end{document}